%% file: ms.tex
\documentclass{article}

\usepackage{microtype}
\usepackage{graphicx}
\usepackage{booktabs} % for professional tables
\usepackage{siunitx} % added by Mohammad for scientific numbers

\usepackage[labelformat=parens]{subcaption}

\usepackage{multirow}
\usepackage{slashbox}
\usepackage{mathtools}

\usepackage{hyperref}

\usepackage[accepted]{icml2020}

\icmltitlerunning{Federated Multi-Mini-Batch: An Efficient Training Approach to Federated Learning in Non-IID Environments}

\begin{document}

\twocolumn[
\icmltitle{Federated Multi-Mini-Batch: An Efficient Training Approach to Federated Learning in Non-IID Environments}



\icmlsetsymbol{equal}{*}

\begin{icmlauthorlist}
\icmlauthor{Reza Nasirigerdeh}{equal,tum1}
\icmlauthor{Mohammad Bakhtiari}{equal,uhh}
\icmlauthor{Reihaneh Torkzadehmahani}{tum1}
\icmlauthor{Amirhossein Bayat}{tum2}
\icmlauthor{Markus List}{tum3}
\icmlauthor{David B. Blumenthal}{fau,jsa}
\icmlauthor{Jan Baumbach}{uhh,jsa}

\end{icmlauthorlist}

\icmlaffiliation{tum1}{AI in Medicine and Healthcare, Technical University of Munich, Munich, Germany}

\icmlaffiliation{tum2}{Department of Informatics, Technical University of Munich,
Munich, Germany}

\icmlaffiliation{tum3}{TUM School of Life Sciences, Technical University of Munich,
Freising, Germany}

\icmlaffiliation{uhh}{Department of Informatics, University of Hamburg,
Hamburg, Germany}

\icmlaffiliation{fau}{Department of Artificial Intelligence in Biomedical Engineering, Friedrich-Alexander University Erlangen-Nürnberg, Erlangen, Germany}

\icmlaffiliation{jsa}{Joint last authors}

\icmlkeywords{Machine Learning, ICML}

\vskip 0.3in
]



\printAffiliationsAndNotice{\icmlEqualContribution} % otherwise use the standard text.

\begin{abstract}
Federated learning has faced performance and network communication challenges, especially in the environments where the data is not independent and identically distributed (IID) across the clients. To address the former challenge, we introduce the \textit{federated-centralized concordance} property and show that the \textit{federated single-mini-batch} training approach can achieve comparable performance as the corresponding centralized training in the \textit{Non-IID} environments. To deal with the latter, we present the \textit{federated multi-mini-batch} approach and illustrate that it can establish a trade-off between the performance and communication efficiency and outperforms \textit{federated averaging} in the \textit{Non-IID} settings.
\end{abstract}

\section{Introduction}
\label{sec:introduction}
\input{introduction}

\section{Preliminaries}
\label{sec:background}
\input{background}

% \section{Related work}
% \label{sec:related_work}
% \input{related_work}

\section{Method}
\label{sec:method}
\input{method}

\section{Results}
\label{sec:result}
\input{result}

\section{Conclusion}
\label{sec:conclusion}
\input{conclusion}

%% Acknowledgements should only appear in the accepted version.
%\section*{Acknowledgements}
%
%\textbf{Do not} include acknowledgements in the initial version of
%the paper submitted for blind review.
%
%If a paper is accepted, the final camera-ready version can (and
%probably should) include acknowledgements. In this case, please
%place such acknowledgements in an unnumbered section at the
%end of the paper. Typically, this will include thanks to reviewers
%who gave useful comments, to colleagues who contributed to the ideas,
%and to funding agencies and corporate sponsors that provided financial
%support.
%

\nocite{langley00}

\bibliography{example_paper}
\bibliographystyle{icml2020}

\end{document}

%% file: introduction.tex
\begin{table*}[!ht]
\centering
\caption{Notations}
\resizebox{0.95\textwidth}{!}{\begin{tabular}{l c c r}
\toprule
\begin{tabular}{l l}
\multicolumn{2}{c}{Common}\\
\cmidrule{1-2}
$i$: & iteration \\
$M$: &model \\
$W_{i}$: & weights of $M$ \\
$S_{i}$: & subset of training samples \\
$F$: & loss function \\
$\nabla$: & gradient \\
$\eta$: & learning rate \\
$D^\prime$: & test set \\
$I_{max}$: & maximum iterations \\
$\delta$: & discordance value \\
$\epsilon$: & concordance threshold \\

\end{tabular}
&
\begin{tabular}{l l}
\multicolumn{2}{c}{Centralized}\\
\cmidrule{1-2}
$D$: & aggregated training set\\
$N$: & sample size of $D$ \\
$B^{\prime}$: & batch size in centralized training \\
$M^{c}$: & centralized model \\
$W^{c}_{i}$: & weights of $M^{c}$ \\
$\ell^{c}_{i}$: & loss value of $M^{c}$ on $D^{\prime}$  \\
\\
\\
\\
\\
\\
\end{tabular}
&
\begin{tabular}{l l}
\multicolumn{2}{c}{Server}\\
\cmidrule{1-2}
$K$: & number of clients \\ 
$B$: &  batch size of clients \\
$C$: &  batch count of clients \\
$E$: & number of local epochs \\ 
$L$: & number of unique labels in clients \\
$M^{g}$: & global (federated) model \\
$W^{g}_{i}$: & weights of $M^{g}$ \\
$\ell^{g}_{i}$: & loss value of $M^{g}$ on $D^{\prime}$\\
\\
\\
\\
\end{tabular}
&
\begin{tabular}{l l}
\multicolumn{2}{c}{Client}\\
\cmidrule{1-2}
$j$: & client index \\
$D_j$: & training set of client \\
$N_j$: & sample size of $D_j$ \\
$w_{ij}^l$: & weights of local model \\
$n^{l}_{ij}$: & number of samples used in training\\
$\mu_j$: & number of local updates\\
\\
\\
\\
\\
\\
\end{tabular}
\\
\bottomrule
\end{tabular}
}
\end{table*}
Federated learning \cite{konevcny2015federated, konevcny2016federated, mcmahan2017communication} is a distributed learning approach that enables multiple parties (clients) to learn a shared (global) model without moving their local data off-site. In federated learning, most of the training is performed by the clients and an aggregation strategy is employed by a central server to iteratively update the global model. The privacy-preserving nature of federated learning has made it popular for applications such as healthcare data analysis \cite{sheller2018multi, brisimi2018federated, chen2020fedhealth} and mobile keyboard prediction \cite{hard2018federated, yang2018applied}, in which access to data is impossible due to strict privacy policies.

Federated averaging (\textit{FedAvg}) \cite{mcmahan2017communication} is a communication-efficient approach to federated learning, which aims to reach an accurate global model with an efficient number of communication rounds between the clients and the server. The main idea behind \textit{FedAvg} is to perform a large number of local updates in the clients and then take a simple weighted average over the local model parameters on the server. \textit{FedAvg} can dramatically reduce the number of communication rounds if the data is independent and identically distributed (\textit{IID}) across the clients. 
 
 However, federated learning faces performance and network communication challenges when it comes to \textit{Non-IID} settings and \textit{FedAvg} as the training approach \cite{zhao2018federated,jeong2018communication, li2019convergence, hsieh2019non,sattler2019robust, li2020federated, wang2020federated,wang2020optimizing, briggs2020federated}. The global model trained by \textit{FedAvg} might not converge to the optimum in \textit{Non-IID} environments, and consequently, federated training might not provide comparable performance as it does for \textit{IID} settings. Moreover, \textit{FedAvg} might still require a large number of communication rounds to achieve target performance in \textit{Non-IID} configurations.

In this paper, we introduce the \textbf{\textit{federated-centralized concordance}} property (Section \ref{sec:method}), which is directly related to the performance challenge in \textit{Non-IID} environments. The property states that the federated (global) model trained by a set of clients on their local data is \textbf{similar to} the centralized trained on the aggregated data. If a federated training approach holds this property, it can achieve comparable performance as the corresponding centralized training regardless of the data and sample distribution across the clients. We experimentally show that the \textbf{\textit{federated single mini-batch} (\textit{FedSMB})} approach (Sections \ref{sec:method}) can train federated models that are concordant with the centralized model, and as a result, it has the potential to tackle the performance challenge in \textit{Non-IID} settings (section \ref{sec:result}).

To address the communication challenge, we present \textbf{\textit{federated multi-mini-batch} (\textit{FedMMB}}) as a generalization of \textit{FedSMB} (Section \ref{sec:method}). The main idea behind \textit{FedMMB} is to decouple the batch size from the batch count and to allow for specifying the number of batches for training the local models at the clients (the number of local updates) independent of the batch size. This decoupling is not possible with \textit{FedAvg}, where a single hyperparameter determines both the batch size and the batch count. Our simulation results illustrate that \textit{FedMMB} can provide a trade-off between the performance and communication efficiency by controlling the number of local updates on the clients (Sections \ref{subsec:result-fed-mmb} and \ref{subsec:resust-vs-fed-avg}). Moreover, \textit{FedMMB} attains higher performance than \textit{FedAvg} in the \textit{Non-IID} environments (Section \ref{subsec:resust-vs-fed-avg}).

% \textit{CIFAR-10}~\cite{cifar10} and \textit{HAM10000}~\cite{ham10000} medical image dataset

%% file: background.tex
\textbf{\textit{Gradient descent}} is the most widely used optimization method for training neural network models. In each iteration $i$, the gradient $\nabla$ of the loss function $F$ of  the model $M$ characterized by the parameters (weights) $W_{i}$ are computed by minimizing $F$ on subset $S_{i}$ of the training samples in the dataset. Then, the model parameters are updated in the opposite direction of the gradient values. The learning rate $\eta$ specifies the step size of the update~\cite{gradient-descent-overview}.
\begin{equation}
    W_{i+1} = W_{i} - \eta  \nabla F(W_{i}; S_{i})
\end{equation}
There are different variants of gradient descent depending on how the samples of the training dataset are employed to update the model parameters. In \textit{full gradient descent} (\textit{FGD}), all samples are leveraged to compute the gradients; \textit{stochastic gradient descent} (\textit{SGD}) calculates the gradients using a single randomly selected sample of the training dataset; \textit{mini-batch gradient descent} (\textit{MBGD}) optimizes the loss function on a random small batch of samples~\cite{gradient-descent-slides,stochastic-gradient-descent}. For large neural networks, trained on very large datasets, \textit{MBGD} is typically the best choice because it is computationally efficient~\cite{gradient-descent-slides}.

A neural network model can be trained in a centralized or distributed (including federated) environment. In centralized training, the whole dataset is located at a single site, and the model is iteratively trained on the dataset using one of the variants of gradient descent. \textit{Epoch} indicates the number of iterations required to employ all samples of the dataset for training. 

\textbf{\textit{Federated learning}} is a privacy-preserving approach to learning a global model from the data distributed across multiple clients. Federated learning can be conducted in a  \textit{cross-device} or \textit{cross-silo} setting~\cite{kairouz2019federated3}. The former involves a huge number of mobile or edge devices as clients, whereas there is a small number of clients (e.g. dozens of medical centers) for training in the latter setting. We assume that the clients have different training samples but the same form of a neural network model; additionally, all clients are selected to participate in the training process in each communication round. 

In each iteration $i$ of the federated training, all $K$ clients obtain the global model parameters $W_{i}^{g}$ from the server and set the weights of their local model to $W_{i}^{g}$. Next, each client $j$ computes the local model parameters $W_{ij}^{l}$ by optimizing the loss function $F$ on $n^{l}_{ij}$ samples from its local data using one of the variants of gradient descent. Afterwards, the server receives the local parameters from the clients and calculates the global model parameters for the next iteration by taking the weighted average over the local parameters:
\begin{equation}
    W_{i+1}^{g} = \frac{\sum^{K}_{j=1}n^{l}_{ij} W_{ij}^{l}}{\sum^{K}_{j=1}n^{l}_{ij}}
\label{eq:fl}
\end{equation}
Each iteration of the federated training updates the global model parameters once and requires one communication round between each client and the server. Therefore, iteration and communication round are used interchangeably in the federated environment. However, the clients might update their local model parameters once or multiple times in each iteration depending on the variant of gradient descent they employ for local optimization. 

\textbf{\textit{FedAvg algorithm}} employs \textit{MBGD} in the clients, aiming to reduce the number of communication rounds by performing more local updates at the clients. In \textit{FedAvg}, each client $j$ updates its local model parameters $\mu_{j} = E \lceil \frac{N_{j}}{B} \rceil$ times, where $E$ is the number of local epochs, $B$ is the batch size, and $N_{j}$ is the number of samples in the training set of client $j$. In other words, the clients run the \textit{MBGD} algorithm $E$ times on the local data before sending the local model parameters to the server. The theoretical analysis on the convergence of \textit{FedAvg} in the \textit{Non-IID} settings shows that \textit{FedAvg} with $E > 1$ and full batch might not converge to the optimum~\cite{li2019convergence}.

\textbf{\textit{Data distribution}} (i.e. feature and label distribution) across the clients can be \textit{IID} or \textit{Non-IID}. In the former, the training sets of the clients have similar (homogeneous) data distributions while in the latter, the data is heterogeneously distributed across the clients. The \textit{sample distribution} among the clients might be \textit{balanced} or \textit{imbalanced}. In the balanced distribution, the sample sizes of the clients are alike, whereas the clients have very different sample sizes in the imbalanced distribution. Hsieh et al. \cite{hsieh2019non} empirically show that data heterogeneity makes accurate federated learning very challenging, and the level of heterogeneity plays a major role in the problem. In this study, we focus on the \textit{Non-IID} label distribution and mainly \textit{balanced} sample distribution.

%% file: method.tex
\begin{algorithm}[tb]
  \caption{\textit{Federated multi-mini-batch}\\
  The server takes $I_{max}$ and $K$ as hyperparameters while $B$, $C$, and $\eta$ are hyperparameters for the clients.
   }
   \label{alg:FedMMB}
\begin{algorithmic}
\STATE \textbf{Server}
\FUNCTION{train:}
    \STATE $W^{g}_{0} \leftarrow$ initialize global weights
   
	\FOR{iteration $i$ from  $0$ to $(I_{max} - 1)$}
		\FOR{client $j$ from $1$ to $K$}
			\STATE $W^{l}_{ij},\  n^{l}_{ij} \leftarrow$ \textbf{$P_{j}.update$}($i$,\ $W^{g}_{i}$)
		\ENDFOR
		\STATE    $W_{i+1}^{g} \leftarrow \frac{\sum^{K}_{j=1}n^{l}_{ij}  W_{ij}^{l}}{\sum^{K}_{j=1}n^{l}_{ij}}$
	\ENDFOR
	\STATE return $W_{I_{max}}^{g}$
\ENDFUNCTION
\\\hrulefill
\STATE \textbf{Client $P_{j}$}
\FUNCTION{update:}
	\STATE $T\leftarrow\lceil \frac{N_{j}}{B}\rceil; \  f \leftarrow\lceil \frac{T}{C}\rceil$
	\STATE $p \leftarrow (i\%f) C;\ q \leftarrow p + C - 1$
	
	\IF{$q >  T - 1$}
	    \STATE $q \leftarrow T - 1$
	\ENDIF
	
	\STATE $n \leftarrow 0; \ u \leftarrow 0; \ W_{0} \leftarrow W^{g}_{i}$
	\FOR{ batch $\beta$ from $\beta_{p}$ to $\beta_{q}$}
	   \STATE $W_{u+1} \leftarrow W_{u} - \eta \nabla F(W_{u}; \beta)$
	   \STATE $u \leftarrow u + 1;\  n \leftarrow n + sizeof(\beta)$
	\ENDFOR
	
	\IF{$(i+1) \% f == 0$}
 		\STATE $\beta_{0} \ ... \  \beta_{(T - 1)} \leftarrow $ shuffle and split $D_{j}$ into batches 
	\ENDIF
	
	\STATE return $W_{u}, \ n$
\ENDFUNCTION
\end{algorithmic}
\end{algorithm}
In this section, we define an \textbf{empirical} property called \textit{federated-centralized concordance}, and describe the \textit{FedSMB} training approach and its generalization, \textit{FedMMB} approach, which can fulfill the performance and network communication challenges in federated learning, respectively.
% Next, we present the \textit{FedMMB} approach as a generalization of \textit{FedSMB} to address the network communication challenge.

\subsection{Federated-centralized concordance}
Consider the federated and centralized settings as follows: The federated setting contains $K$ clients in which each client $j$ possesses training dataset $D_{j}$ with sample size $N_{j}$. In iteration $i$, the clients collaboratively train a federated (global) model $M^{g}$ characterized by weights $W^{g}_{i}$. In the centralized environment, the dataset $D$ with $N$ samples is the same as the aggregation of the training datasets of the clients, i.e. $ D = \sum^{K}_{j=1}D_{j}$ and $N = \sum^{K}_{j=1}N_{j}$. The centralized model $M^{c}$ characterized by weights $W^{c}_{i}$ is iteratively trained on the dataset. $M^{g}$ and $M^{c}$ have the same form and an initialized with the same weights. Both environments employ the same loss function $F$ to optimize the model, and the same learning rate $\eta$ to update the model. The models are evaluated on the test dataset $D^{\prime}$. $\ell_{i}^{g}$ and $\ell_{i}^{c}$ indicate the loss value of $M^{g}$ and $M^{c}$ on $D^{\prime}$ in iteration $i$, respectively.

The \textit{federated-centralized concordance} property: The federated model $M^{g}$ trained on the distributed datasets of the clients ($D_{j}$, $1\le j \le K$) is similar to the centralized model $M^{c}$ trained on the aggregated dataset $D$ if the discordance (dissimilarity) value $\delta$ between the federated and centralized models is less than a very small value $\epsilon$. The discordance value $\delta$ is defined as the mean square error (\textit{MSE}) between the loss values from the federated and centralized models on the test dataset $D^{\prime}$:
\begin{equation}
    \delta = \frac{\sum_{i=1}^{I_{max}}(\ell_{i}^{g} - \ell_{i}^{c})^2}{I_{max}}
\end{equation}
where $I_{max}$ is large enough for both models to converge.

Given that, a federated training approach is \textit{concordant} with a centralized training approach on the dataset \textit{D} if the models trained by the approaches are concordant independent of the data and sample distribution across the clients provided that $ D = \sum^{K}_{j=1}D_{j}$. The practical application of this property is that if a federated approach holds the property, it can provide comparable performance as the corresponding centralized approach even in \textit{Non-IID} environments, and as a result, these environments are not challenging for the federated approach from the performance perspective. 

\textit{FedSMB} is a training approach, where the clients train the model on a single mini-batch from their local dataset instead of the whole in each communication round. In the next section, we experimentally show that the federated models from \textit{FedSMB} with $K$ clients and batch size $B$ are similar to the those from the centralized training using \textit{MBGD} with batch size $B^\prime =B \times K$ under the following assumptions: (1) \textit{FedSMB} and \textit{MBGD} use a relatively small learning rate, (2) the neural network model is convolutional or fully-connected and does not use any regularization such as batch normalization or random dropout, and (3) the sample distribution across the clients is balanced. 

\subsection{FedMMB}
Although \textit{FedSMB} can potentially meet the performance challenge, it suffers from a practical limitation: it is not communication-efficient, requiring a large number of communication rounds to achieve target performance. To tackle this issue, the \textit{FedMMB} approach (Algorithm \ref{alg:FedMMB}) generalizes \textit{FedSMB} by specifying the number of batches (hyperparameter $C$) that clients should employ to locally train the model separate from the batch size (hyperparameter $B$).
 
In the initial step, the server initializes the global model; moreover, each client $j$ shuffles its local dataset of size $N_{j}$ and splits it into $\lceil\frac{N_{j}}{B}\rceil$ batches of size $B$ (except the last one whose size might be less than $B$). In the first iteration, the clients train the global model on the first $C$ batches from their dataset, updating the model parameters $C$ times. Afterwards, each client $j$ sends the updated model as well as the number of samples used for training ($n^{l}_{ij}$) to the server. The server takes the weighted average over the local models from the clients to compute the new global model. Likewise, the clients train the model on the second $C$ batches of their dataset in the second iteration, and the training process is repeated for a pre-specified number of iterations. The client shuffles and splits its dataset every $\lceil\frac{\lceil\frac{N_{j}}{B}\rceil}{C}\rceil$ iteration.

The batch size and the number of batches used to perform local updates in each iteration can dramatically affect the performance and network efficiency in the federated environments (especially \textit{Non-IID} ones). In \textit{FedAvg}, they are coupled to each other because a single hyperparameter (i.e. batch size) determines both. \textit{FedMMB} decouples the batch size from the batch count by using a separate hyperparameter for each of them. This decoupling enables \textit{FedMMB} to control the number of local updates in the clients separate from the batch size. Given that, \textit{FedMMB} can provide a trade-off between the performance and communication efficiency in various \textit{Non-IID} environments (Section \ref{subsec:result-fed-mmb}).
\begin{figure*}
       %A
        \begin{minipage}{\textwidth}
        \centering
        \includegraphics[width=0.95\textwidth]{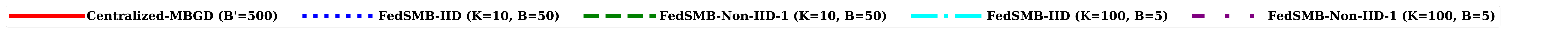}
    \end{minipage}
    \begin{minipage}{.45\textwidth}
        \centering
        \includegraphics[width=1\textwidth]{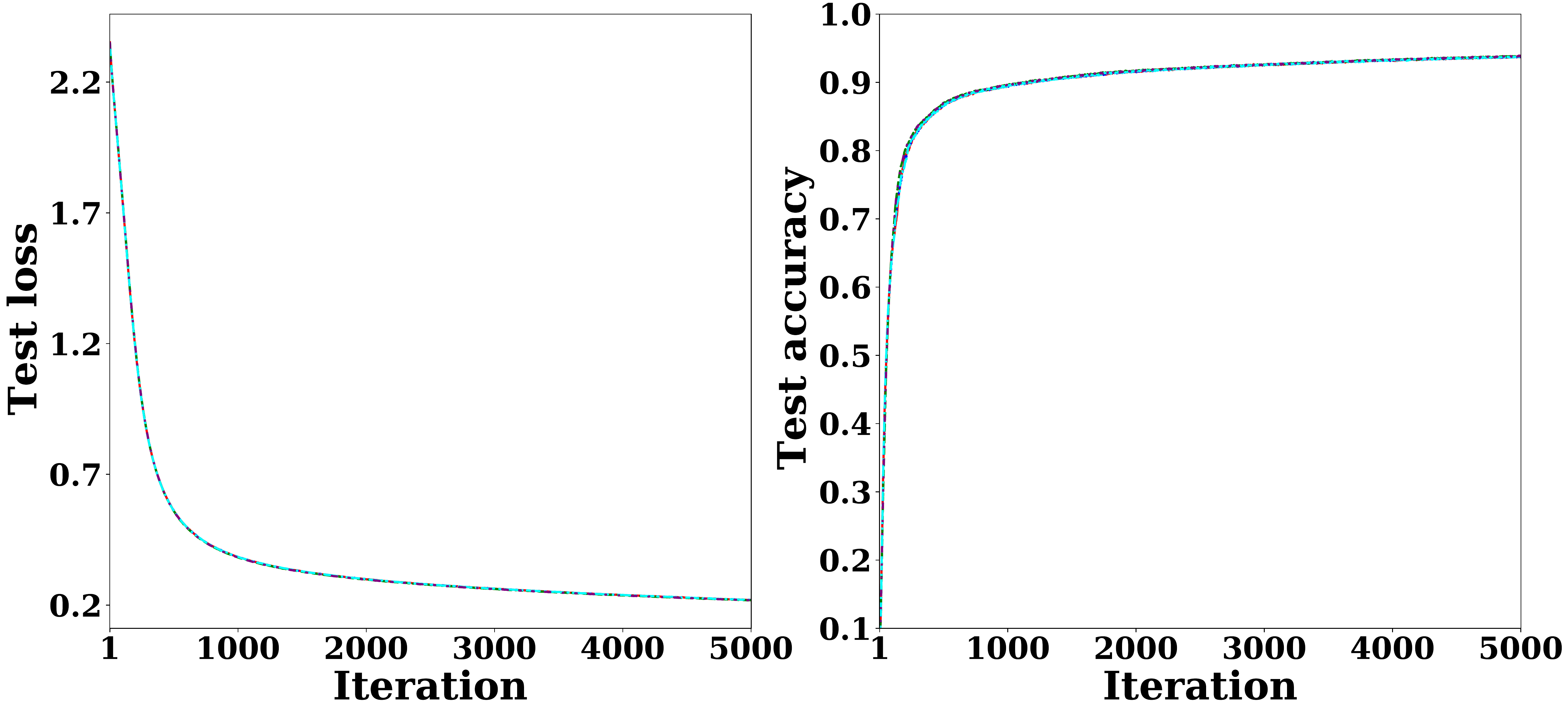}
        \subcaption{\textit{2FNN-MNIST}}  
    \end{minipage}
    \hfill
    \begin{minipage}{.45\textwidth}
        \centering
        \includegraphics[width=1\textwidth]{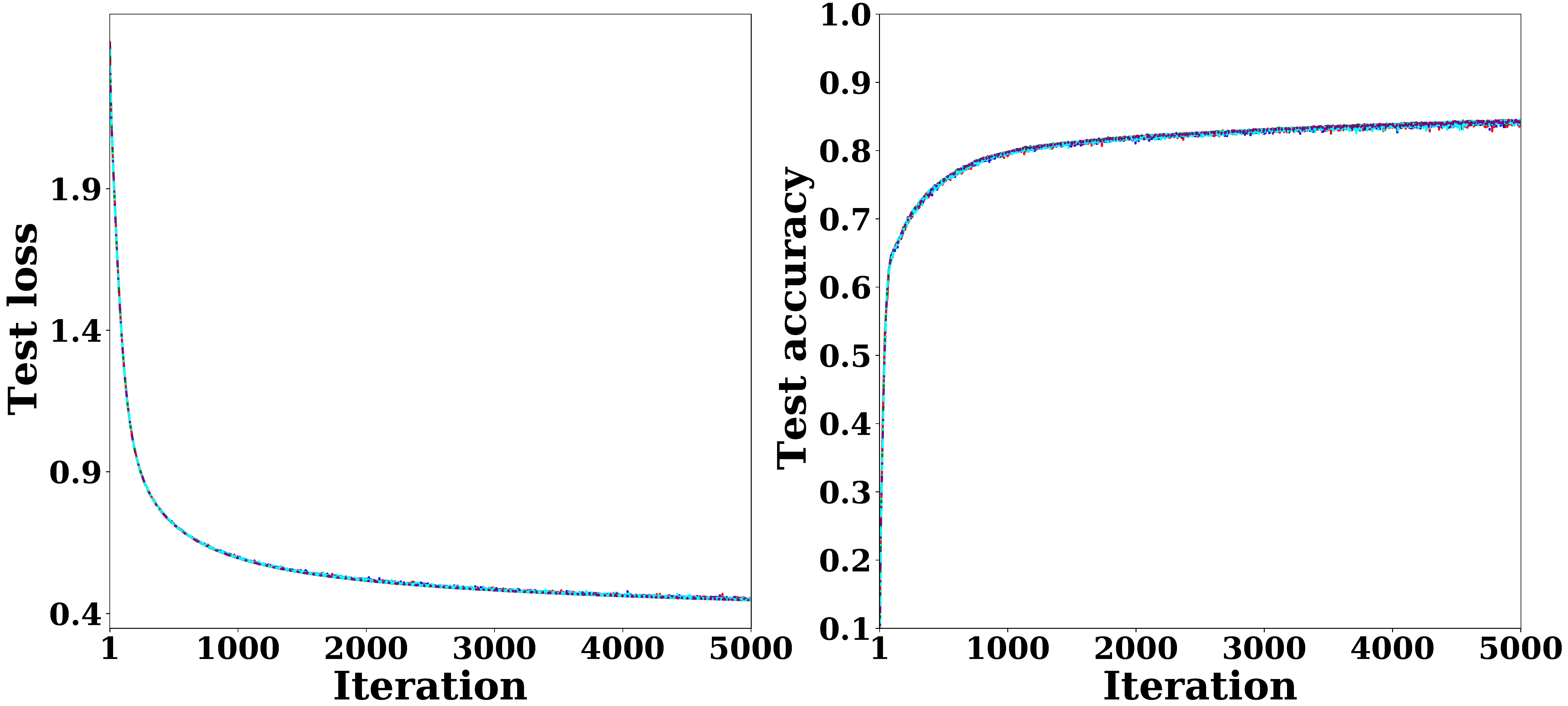}
        \subcaption{\textit{2FNN-FMNIST}}  
    \end{minipage}
        \begin{minipage}{.45\textwidth}
        \centering
        \includegraphics[width=1\textwidth]{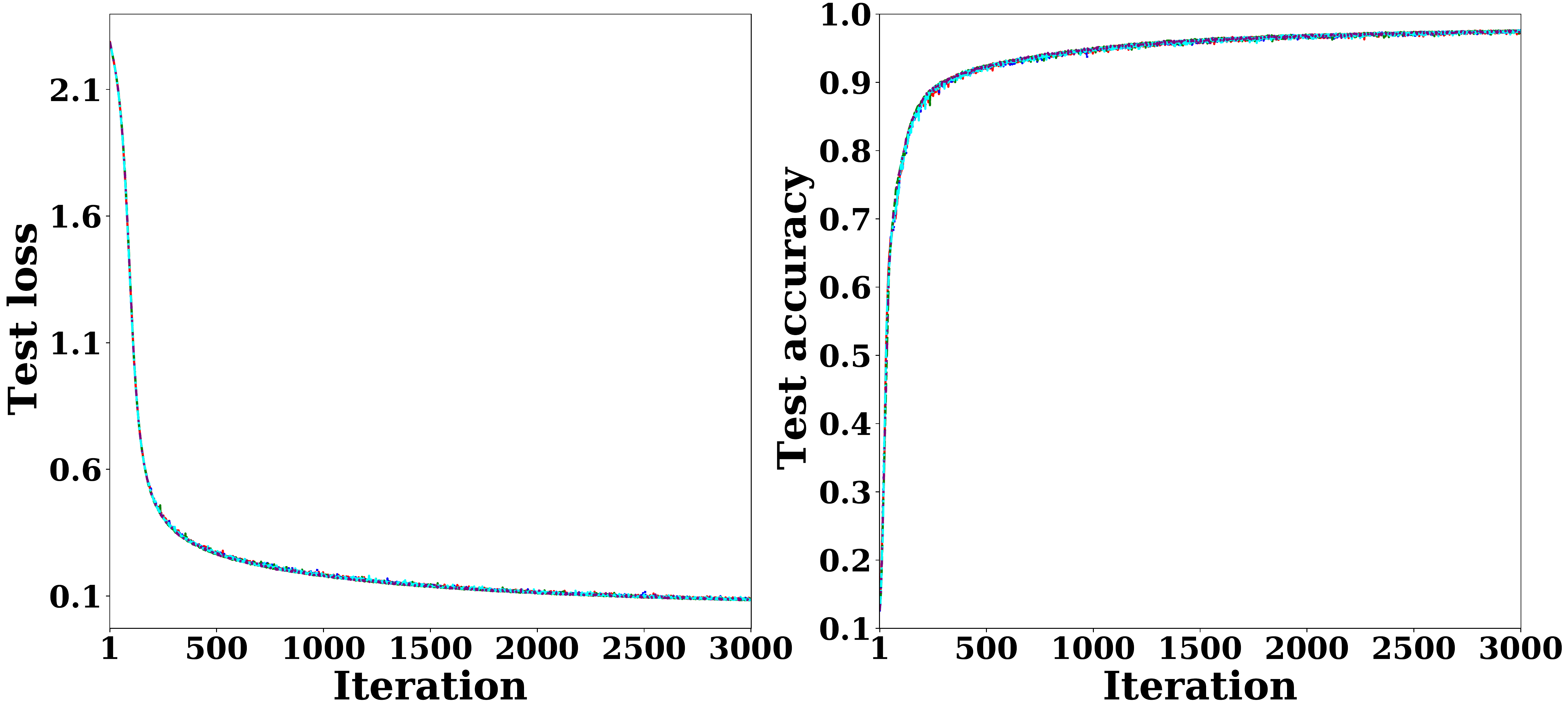}
        \subcaption{\textit{3CFNN-MNIST}}  
    \end{minipage}
    \hfill
    \begin{minipage}{.45\textwidth}
        \centering
        \includegraphics[width=1\textwidth]{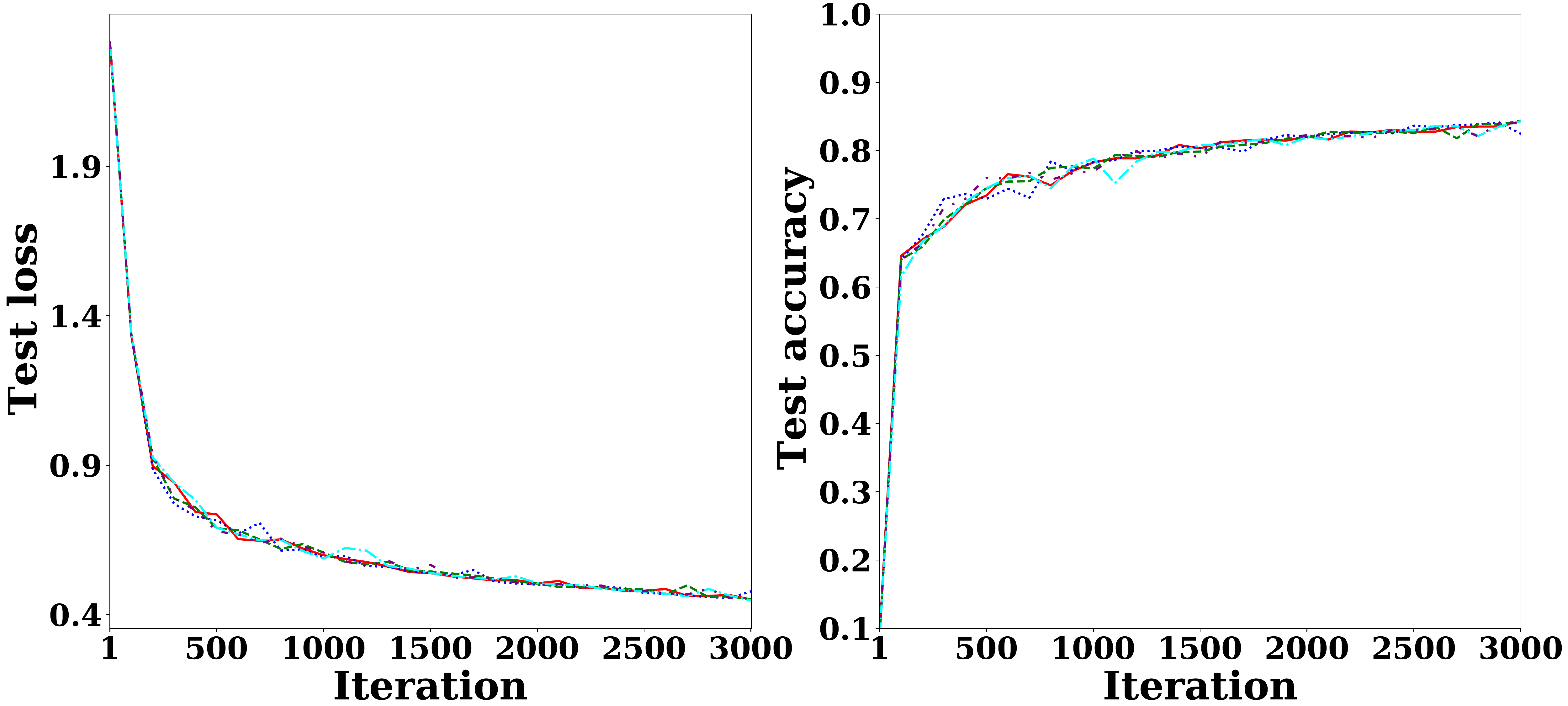}
        \subcaption{\textit{3CFNN-FMNIST}}  
    \end{minipage}
    \caption{Similarity between the federated models from \textit{FedSMB} and those from the centralized training with \textit{MBGD} ($\eta=0.01$).}
    \label{fig:fed-smb-mnist}
\end{figure*}

\begin{table*}[!ht]
\begin{center}
\caption{Discordance $\delta$ $\vert$ accuracy values associated with the scenarios from Figure \ref{fig:fed-smb-mnist}}
\label{tab:fed-smb-mnist}
\resizebox{0.9\textwidth}{!}
{
\begin{tabular}{l c c c r}
Scenario & 2FNN-MNIST & 2FNN-FMNIST & 3CFNN-MNIST & 3CFNN-FMNIST \\
\toprule
\textit{Centralized ($B^{\prime}$=$500$)} 
                & $\ \ \ \ \ \ -\ \ \ \ \ \ $ $\vert\hphantom{.}$ $0.9382$ 
				&  $\ \ \ \ \ \ -\ \ \ \ \ \ $ $\vert\hphantom{00}$ $0.8449$ 
				&   $\ \ \ \ \ \ -\ \ \ \ \ \ $ $\hphantom{1}\vert$ $0.9760$ 
				&   $\ \ \ \ \ \ -\ \ \ \ \ \ $ $\hphantom{1}\vert$ $0.8441$ \\[0.1cm]
\textit{IID ($K$=$10$, $B$=$50$)} 
                & $3 \times 10^{-7}$ $\vert\hphantom{.}$ $0.9383$ 
				&  $5 \times 10^{-6}$ $\vert\hphantom{00}$ $0.8449$ 
				&   $1 \times 10^{-5}$ $\hphantom{1}\vert$ $0.9765$ 
				&   $6 \times 10^{-4}$ $\hphantom{1}\vert$ $0.8443$ \\[0.1cm]
\textit{Non-IID-1 ($K$=$10$, $B$=$50$)} 
                & $3 \times 10^{-6}$ $\vert\hphantom{.}$ $0.9390$ 
				&  $6 \times 10^{-6}$ $\vert\hphantom{00}$ $0.8452$ 
				&   $2 \times 10^{-5}$ $\hphantom{1}\vert$ $0.9764$ 
				&   $5 \times 10^{-4}$ $\hphantom{1}\vert$ $0.8444$ \\[0.1cm]
\textit{IID ($K$=$100$, $B$=$5$)} 
                & $3 \times 10^{-7}$ $\vert\hphantom{.}$ $0.9377$ 
				&  $5 \times 10^{-6}$ $\vert\hphantom{00}$ $0.8451$ 
				&   $1 \times 10^{-5}$ $\hphantom{1}\vert$ $0.9761$ 
				&   $7 \times 10^{-4}$ $\hphantom{1}\vert$ $0.8448$ \\[0.1cm]
\textit{Non-IID-1 ($K$=$100$, $B$=$5$)}
                & $4 \times 10^{-6}$ $\vert\hphantom{.}$ $0.9393$ 
				&  $6 \times 10^{-6}$ $\vert\hphantom{00}$ $0.8452$ 
				&   $2 \times 10^{-5}$ $\hphantom{1}\vert$ $0.9763$ 
				&   $5 \times 10^{-4}$ $\hphantom{1}\vert$ $0.8442$\\[0.1cm]
\bottomrule
\end{tabular}}

\end{center}

\end{table*}
\begin{figure*}[!ht]
\centering
\begin{minipage}[t]{0.6\textwidth}
\centering
\begin{minipage}{0.49\textwidth}
\centering
        \includegraphics[width=1\textwidth]{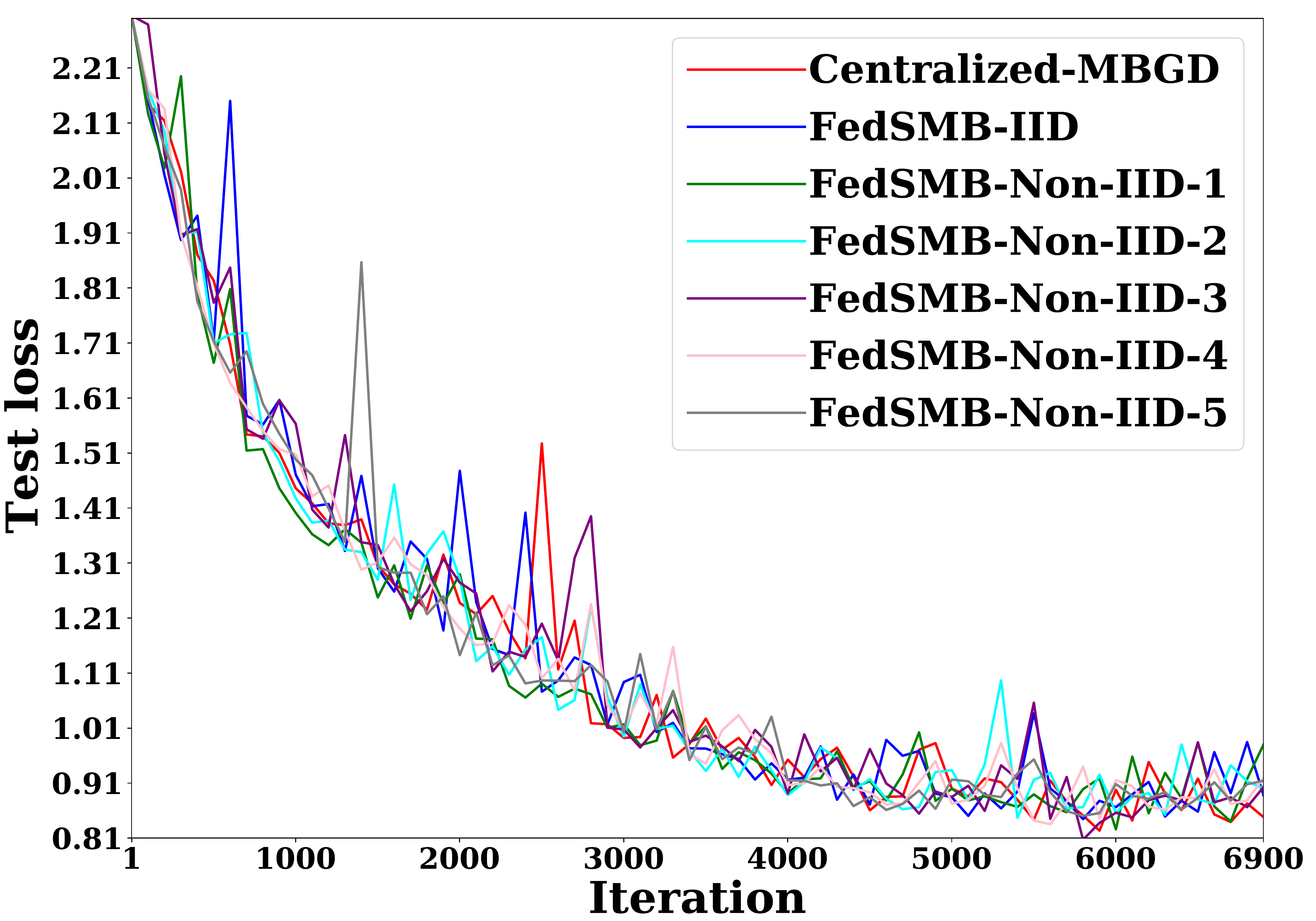}

\end{minipage}
\begin{minipage}{0.49\textwidth}
\centering
        \includegraphics[width=1\textwidth]{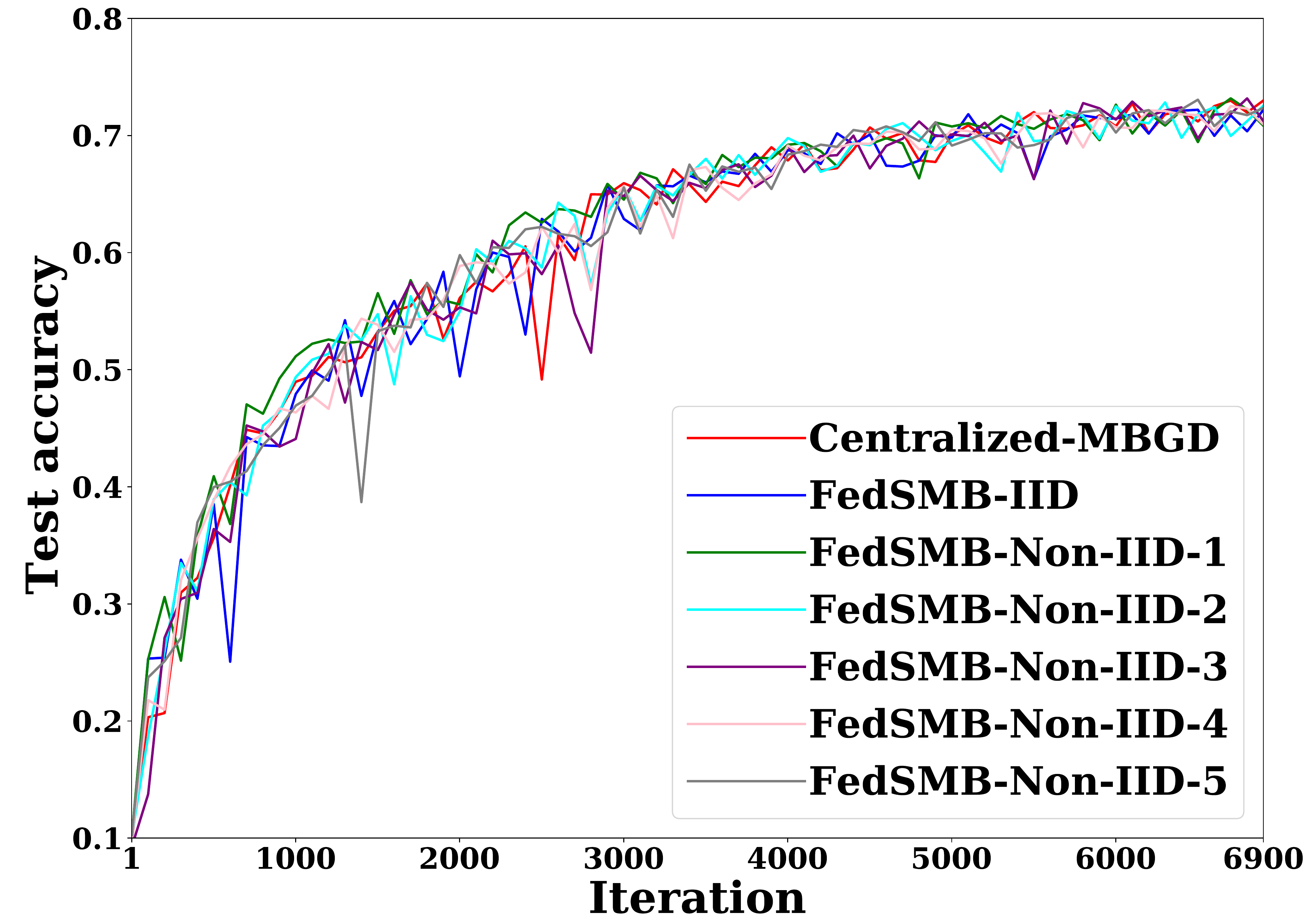}
\end{minipage}
\captionof{figure}{\textit{FedSMB} training for the \textit{4CFNN} model on the CIFAR-10 dataset ($\eta=0.08$, $K=10$, $B=10$, $B^{\prime} = 100$).}
\label{fig:fed-smb-cifar}
\end{minipage}
\hfill
\begin{minipage}{0.33\textwidth}
\centering
\captionof{table}{Discordance $\delta$ and accuracy values corresponding to the scenarios from Figure \ref{fig:fed-smb-cifar}}
\label{tab:fed-smb-cifar}
\resizebox{\textwidth}{!}
{
\begin{tabular}{l l r}
\multicolumn{3}{c}{FedSMB on 4CFNN-CIFAR-10}\\
\hline
Scenario & Discordance  & Accuracy \\[0.1cm]
 \toprule
 
\textit{Centralized} & $\ \ \ \ \ \ -\ \ \ \ \ \ $& $0.7373$ \\[0.1cm]
\textit{IID} & $\num{6e-3}$& $0.7345$ \\[0.1cm]
\textit{Non-IID-1}& $\num{7e-3}$& $0.7350$\\[0.1cm]
\textit{Non-IID-2}& $\num{7e-3}$& $0.7357$\\[0.1cm]
\textit{Non-IID-3} & $\num{7e-3}$ & $0.7390$ \\[0.1cm]
\textit{Non-IID-4} & $\num{6e-3}$ & $0.7342$ \\[0.1cm]
\textit{Non-IID-5} & $\num{7e-3}$ & $0.7358$ \\[0.1cm]
\bottomrule
\end{tabular}
}
\end{minipage}
\end{figure*}

%% file: result.tex
\begin{figure*}[!ht]
       %A
        \begin{minipage}{\textwidth}
        \centering
        \includegraphics[width=0.6\textwidth]{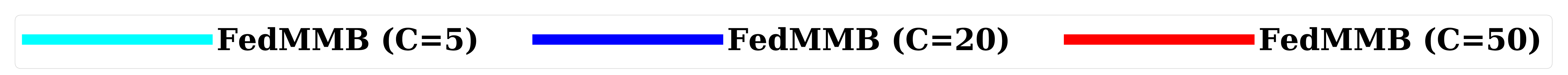}
    \end{minipage}
    \centering
    \begin{minipage}{.25\textwidth}
        \centering
        \includegraphics[width=1\textwidth]{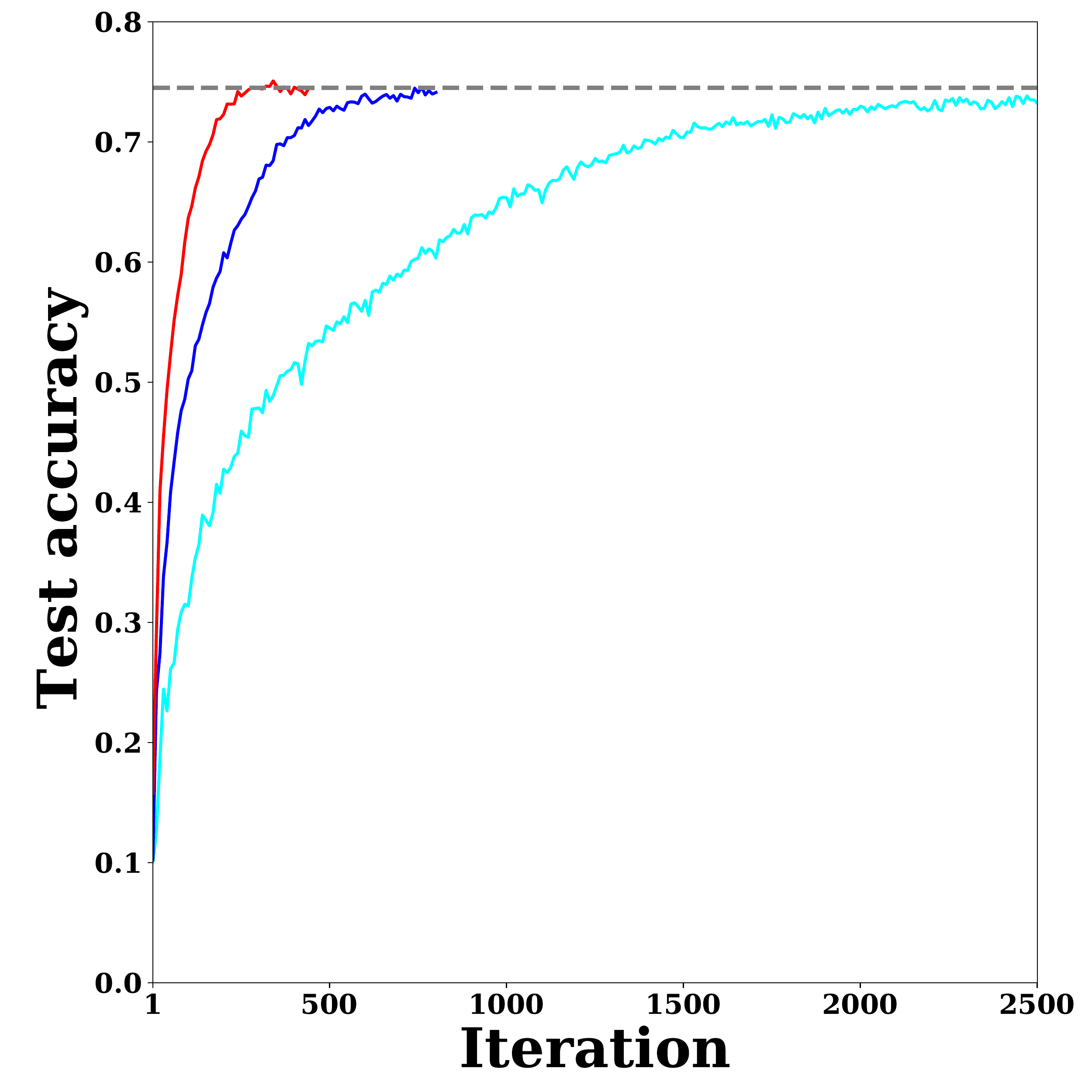}
        \subcaption{\textit{IID}}  
    \end{minipage}
    \begin{minipage}{.25\textwidth}
        \centering
        \includegraphics[width=1\textwidth]{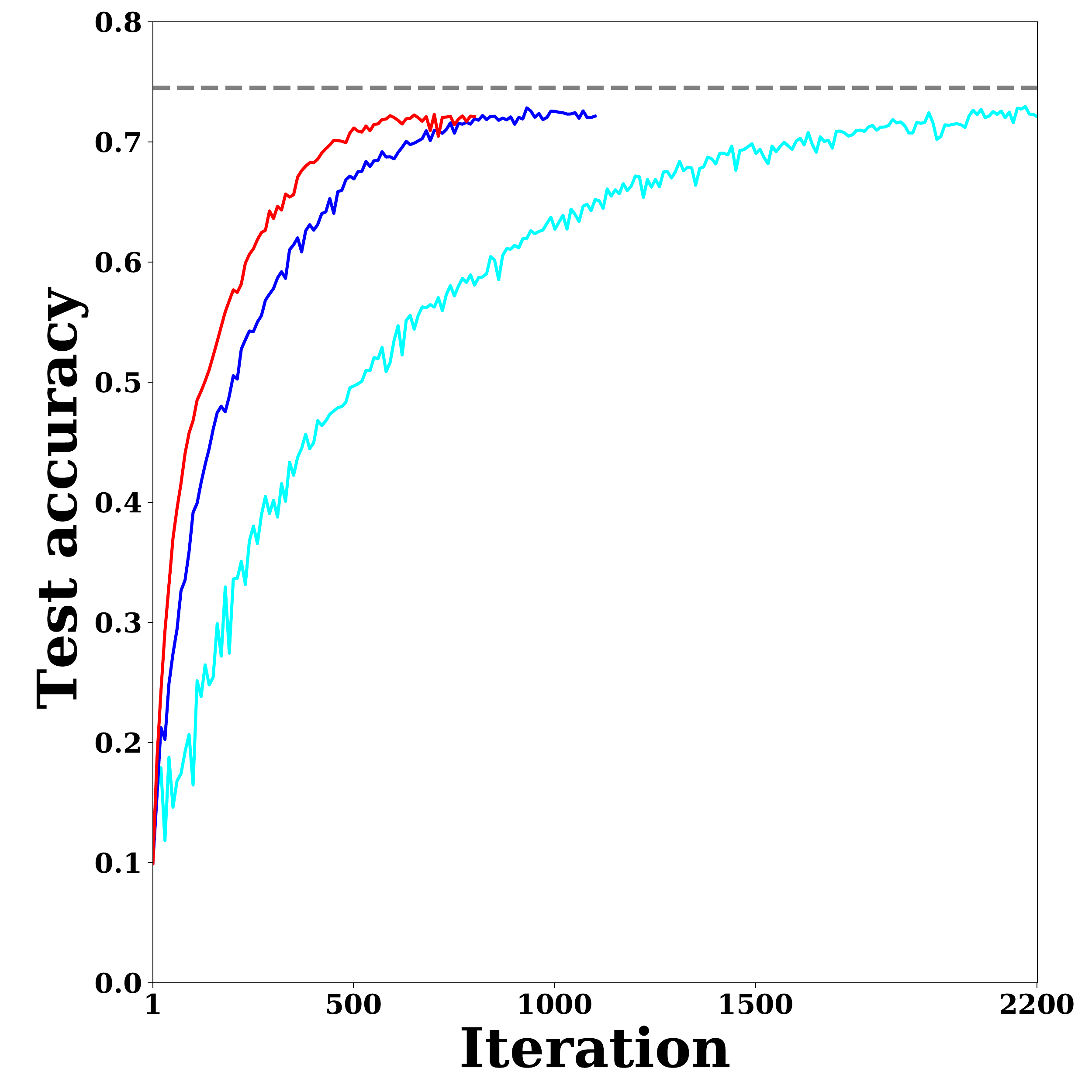}
        \subcaption{\textit{Non-IID-4}}  
    \end{minipage}
    \begin{minipage}{.25\textwidth}
        \centering
        \includegraphics[width=1\textwidth]{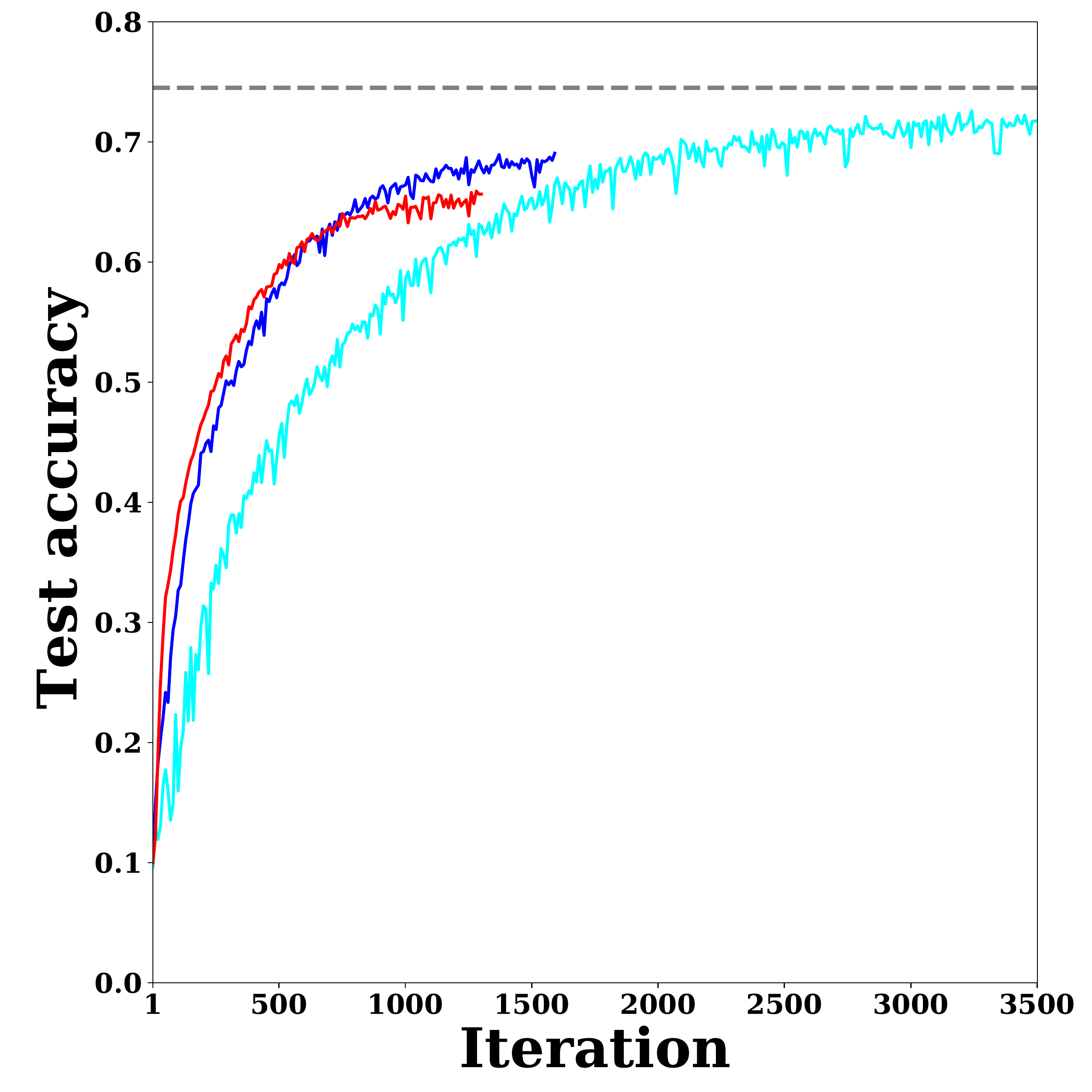}
        \subcaption{\textit{Non-IID-2}}  
    \end{minipage}
     \caption{\textit{FedMMB} training for \textit{4CFNN} on CIFAR-10:  Higher batch counts improve the communication efficiency. In the \textit{Non-IID} environments, \textit{FedMMB} provides comparable accuracy, especially with lower batch counts. In the \textit{Non-IID-2} scenario, larger batch counts (more local updates) adversely affect the model performance. $B=10$ and $K=10$ in all scenarios.  $\eta=0.05$ for the \textit{IID} case; $\eta=0.08,0.05,0.02$ for $C=5,20,50$ in the \textit{Non-IID} settings, respectively. The dashed line indicates the baseline accuracy.}
    \label{fig:fed-mmb}
\end{figure*}
\begin{table}[!ht]
\centering
\caption{Communication rounds $\vert$ maximum accuracy associated with the scenarios in Figure \ref{fig:fed-mmb}.}
\resizebox{\columnwidth}{!}{
\begin{tabular}{l c c c}
\multicolumn{4}{c}{FedMMB on 4CFNN-CIFAR-10}\\
\hline
 & C=5 & C=20 & C=50 \\[0.1cm]
 \toprule
\textit{IID} & 2471 $\vert$ $0.7383$ & $761\ \ $ $\vert$ $0.7456$ & $341\ \ $ $\vert$ $0.7508$ \\[0.1cm]
\textit{Non-IID-4} & $2171$ $\vert$ $0.7295$ & $931\ \ $ $\vert$ $0.7284$ & $701\ \ $ $\vert$ $0.7230$ \\[0.1cm]
\textit{Non-IID-2} & $3241$ $\vert$ $0.7260$ & $1591$ $\vert$ $0.6906$ & $1281$ $\vert$ $0.6591$ \\[0.1cm]
\bottomrule
\end{tabular}
}
\label{tab:fed-mmb}
\end{table}
We first show that the \textit{FedSMB} can train models that are concordant with the centralized \textit{MBGD} models considering the underlying assumptions (e.g. small learning rates or balanced sample distribution). To this end, we leverage the MNIST~\cite{mnist} and Fashion-MNIST (FMNIST)~\cite{fashion-mnist} as datasets, which include 70000 gray-scale images (60000 for training and 10000 for testing) of shape 28x28 as well as 10 label values. Following~\cite{mcmahan2017communication}, we train two different neural network models\footnote{All models are implemented in TensorFlow/Keras~\cite{tensorflow,keras} and use \textit{SGD} optimizer and categorical cross-entropy loss function.} on the datasets: (1) a fully-connected neural network with two hidden layers of size 200 and (2) a convolutional neural network containing two 5x5 convolutional layers, each followed by a 2x2 max-pooling layer. The convolutional layers have 32 and 64 filters, respectively. The second max-pooling layer is followed by a fully-connected layer of size 512. In the models, the fully-connected layers use ReLU while the output layer utilizes the softmax activation function. We refer to the models as \textbf{2FNN} and \textbf{3CFNN}, respectively.

We also evaluate \textit{FedMMB} (and \textit{FedSMB} as its special case) using a more complex model and the CIFAR-10 dataset~\cite{cifar10}. The CIFAR-10 dataset contains 60000 color images (50000 train and 10000 test samples) of shape 32x32 and 10 class labels. We augment the train images by randomly flipping left/right and adjusting the brightness, contrast, saturation, and hue. The train size is doubled after augmentation. The model consists of three 3x3 convolutional layers with 128, 256, and 512 filters, respectively. Each convolutional layer is followed by a 2x2 max-pooling layer. The third max-pooling layer is followed by a fully-connected layer of size 1024. The convolutional and fully-connected layers employ ReLU wheres the output layer has softmax as the activation function. We call this model \textbf{4CFNN}. 

To compare \textit{FedMMB} with \textit{FedAvg}, we employ \textit{4CFNN} and CIFAR-10 as well as the \textit{VGG16} model~\cite{vgg16} and the HAM10000 dataset~\cite{ham10000}. HAM10000 is an imbalanced dataset, comprising $10015$ dermatoscopic skin lesion images of seven classes: \textit{Melanocytic nevi} (6705), \textit{Melanoma} (1113), \textit{Benign keratosis} (1099), \textit{Basal cell carcinoma} (514), \textit{Actinic keratoses} (327), \textit{Vascular lesions} (142), and \textit{Dermatofibroma} (115)\footnote{The numbers inside parentheses indicate the number of samples from each class}. The original resolution of the images is 600x450 but we downsampled them to 200x150  to reduce the number of model parameters. \textit{VGG16} is a deep neural network model containing $13$ convolutional and two fully-connected layers (\textit{TensorFlow} implementation). The model contains $\approx82$ million trainable parameters in our case. 

We distribute the MNIST, FMNIST, and CIFAR-10 datasets across the clients in two different ways: \textit{IID} and \textit{Non-IID}. In the former, the distribution of the label values is similar among the clients, and each client has samples from all ten labels. In the latter, the clients have heterogeneous label distributions. For the \textit{IID} case, we first shuffle the dataset, and then split it into $K$ partitions with the same sample size, and give each partition to one of the $K$ clients. In the \textit{Non-IID} configuration, we have parameter $L$, which indicates the number of unique labels per client and determines the level of the label distribution heterogeneity across the clients. For instance, $L=2$ results in a \textit{Non-IID} setting, where each client only contains the samples from two labels. For a \textit{Non-IID} scenario, we group the samples according to their labels. Next, we divide each group into $(K \times L)/10$ partitions and allocate $L$ partitions with different labels to a client. We assume that the number of clients is divisible by 10. Notice that the sample distribution across the clients is balanced in all scenarios. We refer to a \textit{Non-IID} scenario with parameter $L$ as \textit{Non-IID-L} (e.g. \textit{Non-IID-1}, and \textit{Non-IID-2}). We describe the distribution of the HAM10000 dataset among the clients in section \ref{subsec:resust-vs-fed-avg}. 

\subsection{FedSMB}
\label{subsec:result-fed-smb}
To illustrate the similarity between \textit{FedSMB} and centralized \textit{MBGD} models, we train \textit{2FNN} and \textit{3CFNN} on the MNIST and FMNIST datasets (Figure \ref{fig:fed-smb-mnist} and Table \ref{tab:fed-smb-mnist}) as well as the \textit{4CFNN} model on the CIFAR-10 dataset (Figure \ref{fig:fed-smb-cifar} and Table \ref{tab:fed-smb-cifar}). The \textit{2FNN}, \textit{3CFNN}, and \textit{4CFNN} models are trained in the centralized environment using \textit{MBGD} with $B^\prime=500$, $B^\prime=500$, and $B^\prime=100$, respectively. In the federated environment, \textit{2FNN} and \textit{3CFNN} employ \textit{FedSMB} with $B=50$ and $K=10$ clients, and  $B=5$ and $K=100$ clients under \textit{IID} and \textit{Non-IID-1} settings while \textit{4CFNN} leverages \textit{FedSMB} with $B=10$ , $K=10$ clients under the \textit{IID} and \textit{Non-IID-1} to \textit{Non-IID-5} configurations. The learning rates are $0.01$, $0.01$, and $0.08$ for the models, respectively.

According to Figures \ref{fig:fed-smb-mnist} and \ref{fig:fed-smb-cifar}, the loss and accuracy curves for the centralized and federated models are similar to each other; additionally, \textit{FedSMB} can reach the accuracy of the centralized training regardless of the label distribution among the clients (Tables \ref{tab:fed-smb-mnist} and \ref{tab:fed-smb-cifar}). However, it might need a large number of communication rounds to this end even in the \textit{IID} setting, which implies \textit{FedSMB} is not a communication-efficient approach (Figure \ref{fig:fed-smb-cifar}).

We also compute the discordance value $\delta$ between the federated and centralized models for each federated scenario (Tables \ref{tab:fed-smb-mnist} and \ref{tab:fed-smb-cifar}). We consider $\epsilon=0.01$ as the concordance threshold, i.e. the federated model is concordant with the centralized model if $\delta$ is less than $0.01$. We observe that the discordance $\delta$ between the federated and centralized model is $\num{7e-3}$ in the worst case (the higher discordance value in \textit{4CFNN}-CIFAR-10 is partly due to the higher learning rate used to train the models). These results indicate that the federated training with $K$ clients and batch size $B$ using \textit{FedSMB} and the centralized training with batch size $B^\prime = B \times K$ using \textit{MBGD} provide concordant models.
\subsection{FedMMB}
\label{subsec:result-fed-mmb}
To investigate the efficiency of \textit{FedMMB}, we employ a setting similar to the \textit{FedSMB} case using the \textit{4CFNN} model, the CIFAR-10 dataset, $10$ clients with batch size of $10$, and the best accuracy from the centralized training ($0.7456$) as the baseline. We train the model using different values of $C$ (batch count) under the \textit{IID}, \textit{Non-IID-2} (\textit{severely} \textit{Non-IID} label distribution), and \textit{Non-IID-4} (\textit{moderately} \textit{Non-IID} label distribution) scenarios (Figure \ref{fig:fed-mmb} and Table \ref{tab:fed-mmb}). 

In the \textit{IID} configuration, \textit{FedMMB} can achieve the accuracy of the baseline using high batch count values (C=$20$, $50$). Additionally, the larger batch count ($C$=$50$) requires fewer communication rounds to this end. Thus, increasing the batch count of \textit{FedMMB} in the \textit{IID} environment makes the approach more communication-efficient without compromising the accuracy.

\begin{figure*}[!ht]
    \centering
    \begin{minipage}{.45\textwidth}
        \centering
        \includegraphics[width=1\textwidth]{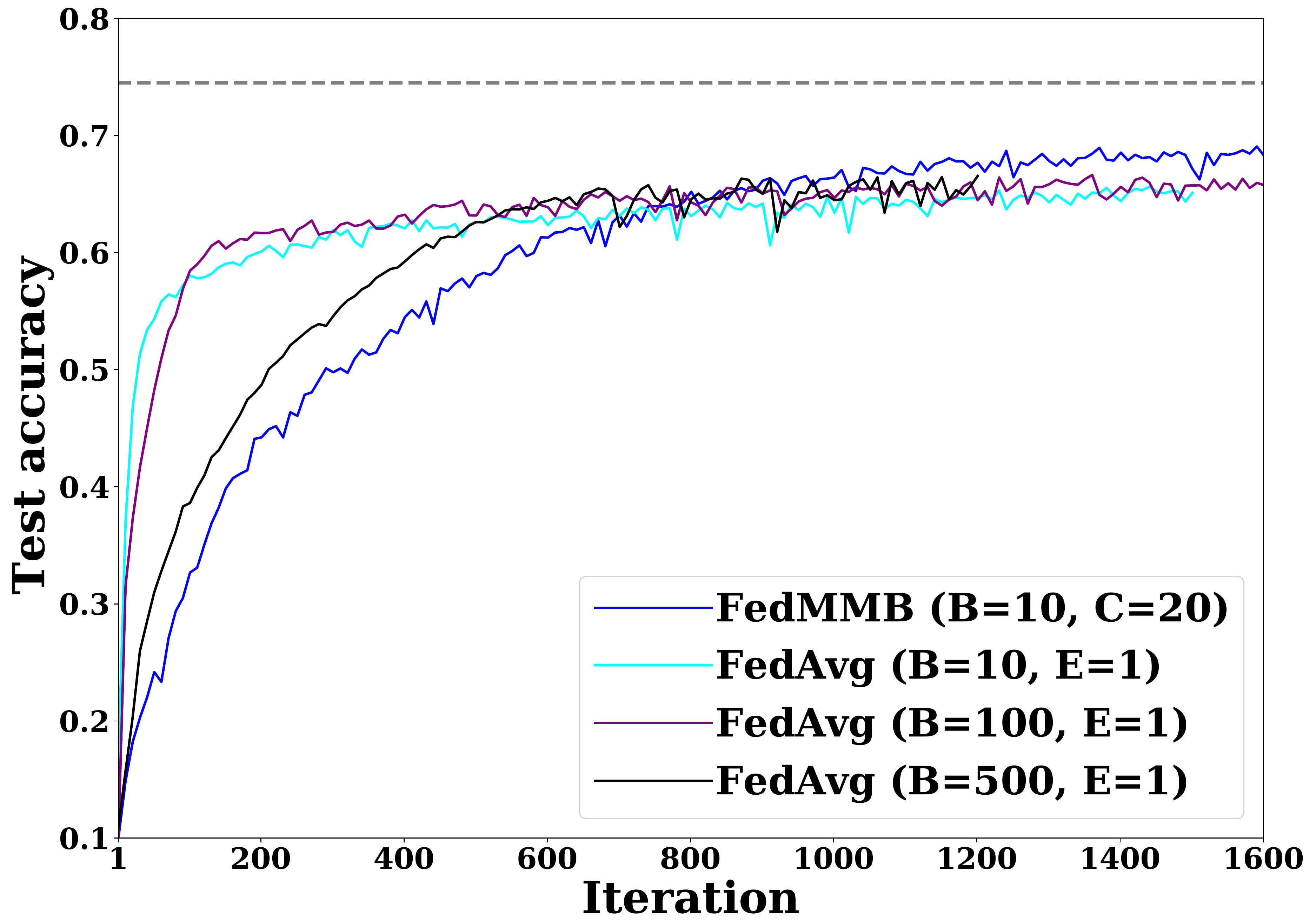}
        \subcaption{\textit{\textit{4CFNN}} on CIFAR-10 under \textit{Non-IID-2} scenario}  
    \end{minipage}
    \hfill
    \begin{minipage}{.45\textwidth}
        \centering
        \includegraphics[width=1\textwidth]{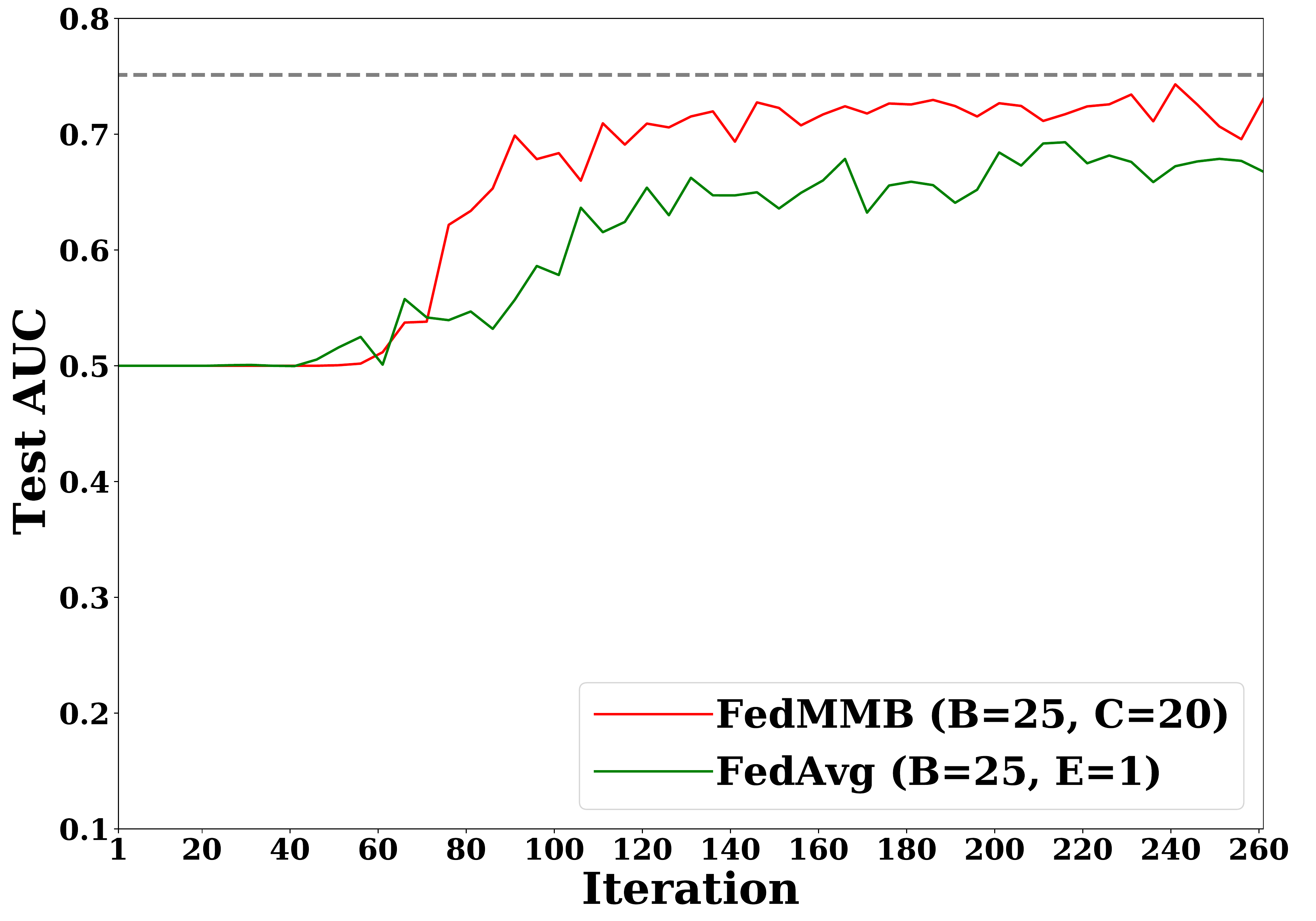}
        \subcaption{\textit{VGG16} on HAM10000 under \textit{HAM-Non-IID} scenario}  
    \end{minipage}
     \caption{Comparison between \textit{FedMMB} and \textit{FedAvg}: \textit{FedMMB} outperforms \textit{FedAvg} in terms of accuracy (a) and \textit{AUC} (b) on the \textit{4CFNN}-CIFAR-10 and \textit{VGG16}-HAM10000 model-dataset pairs, respectively. The dashed line indicates the baseline accuracy or \textit{AUC}. In (a), $\eta=0.02$ for \textit{FedAvg with $B=10$} and $\eta=0.05$ for the other scenarios; $K=10$ for all scenarios. In (b), $K=3$ and $\eta=0.001$ for all scenario.}
    \label{fig:fedmmb-vs-fedavg}
\end{figure*}

\begin{table*}[!htb]
    \caption{Communication rounds and maximum accuracy or \textit{AUC} corresponding to the scenarios from Figure \ref{fig:fedmmb-vs-fedavg}}
\label{tab:fedmmb-vs-fedavg}

    \begin{subtable}{.49\linewidth}
      \centering
        \caption{4CFNN-CIFAR-10}
        \label{tab:cifar-fedmmb-vs-fedavg}
            \resizebox{0.95\columnwidth}{!}{
            \begin{tabular}{l c c }
        \hline
         & Communication rounds & Accuracy \\[0.1cm]
         \toprule
        \textit{\textbf{FedMMB} ($B$=$10$, $C$=$20$)} & $1591$ &  $\mathbf{0.6906}$ \\[0.1cm]
        \textit{FedAvg ($B$=$10$, $E$=$1$)} & $1441$ & $0.6564$\\[0.1cm]
        \textit{FedAvg ($B$=$100$, $E$=$1$)} & $1361$ & $0.6663$\\[0.1cm]
        \textit{FedAvg ($B$=$500$, $E$=$1$)} & $1201$ & $0.6654$ \\[0.1cm]
        \bottomrule
    \end{tabular}
    }
    \end{subtable}%
    \hfill
    \begin{subtable}{.49\linewidth}
      \centering
        \caption{VGG16-HAM10000}
        \label{tab:ham-fedmmb-vs-fedavg}
            \resizebox{0.95\columnwidth}{!}{\begin{tabular}{l c c }
        \hline
         & Communication rounds & AUC \\[0.1cm]
         \toprule
        \textit{\textbf{FedMMB} ($B$=$25$, $C$=$20$)} & $241$ &  $\mathbf{0.7431}$ \\[0.1cm]
        \textit{FedAvg ($B$=$25$, $E$=$1$)} & $216$ & $0.6931$\\[0.1cm]
        \\
        \\
        \bottomrule
    \end{tabular}
    }
    \end{subtable} 
\end{table*}

% \begin{table}[!ht]
% \centering
% \caption{Communication rounds $\vert$ accuracy corresponding to the scenarios in Figure \ref{fig:fedmmb-vs-fedavg}}
% \label{tab:fedmmb-vs-fedavg}

% \begin{minipage}{.35\textwidth}
%     \resizebox{\columnwidth}{!}{\begin{tabular}{l c c }
%         \multicolumn{3}{c}{FedMMB vs. FedAvg on 4CFNN-CIFAR-10}\\
%         \hline
%          & Communication rounds & Accuracy \\[0.1cm]
%          \toprule
%         \textit{\textbf{FedMMB} ($B$=$10$, $C$=$20$)} & $1591$ &  $\mathbf{0.6906}$ \\[0.1cm]
%         \textit{FedAvg ($B$=$10$, $E$=$1$)} & $1441$ & $0.6564$\\[0.1cm]
%         \textit{FedAvg ($B$=$100$, $E$=$1$)} & $1361$ & $0.6663$\\[0.1cm]
%         \textit{FedAvg ($B$=$500$, $E$=$1$)} & $1201$ & $0.6654$ \\[0.1cm]
%         \bottomrule
%     \end{tabular}
%     }
% \end{minipage}
% \begin{minipage}{.35\textwidth}
%     \resizebox{\columnwidth}{!}{\begin{tabular}{l c c }
%         \multicolumn{3}{c}{FedMMB vs. FedAvg on 4CFNN-CIFAR-10}\\
%         \hline
%          & Communication rounds & Accuracy \\[0.1cm]
%          \toprule
%         \textit{\textbf{FedMMB} ($B$=$10$, $C$=$20$)} & $1591$ &  $\mathbf{0.6906}$ \\[0.1cm]
%         \textit{FedAvg ($B$=$10$, $E$=$1$)} & $1441$ & $0.6564$\\[0.1cm]
%         \textit{FedAvg ($B$=$100$, $E$=$1$)} & $1361$ & $0.6663$\\[0.1cm]
%         \textit{FedAvg ($B$=$500$, $E$=$1$)} & $1201$ & $0.6654$ \\[0.1cm]
%         \bottomrule
%     \end{tabular}
%     }
% \end{minipage}

% \end{table}

In the \textit{Non-IID} scenarios, \textit{FedMMB} never reaches the baseline accuracy. In the moderately \textit{Non-IID} label distribution scenario, all three batch count values achieve a similar accuracy ($0.7295$, $0.7284$, $0.7230$ for $C=5,20,50$), and higher batch counts need fewer communication rounds to this end. In the severely \textit{Non-IID} label distribution case, lower batch counts achieve better accuracy ($0.7260$ vs. $0.6906$ vs. $0.6591$) but with more network communication overhead.

In summary, \textit{FedMMB} with large $C$ values is a realistic choice for the \textit{IID} environment because it can save a huge number of communication rounds without negatively affecting the accuracy. For the \textit{Non-IID} environments, \textit{FedMMB} can establish a trade-off between the accuracy and communication efficiency through the batch count hyperparameter. In scenarios where the accuracy has priority over the communication efficiency, smaller batch count values can be used. Otherwise, a larger batch count is a better choice because it can considerably reduce the network communication overhead. In general, the best value of $C$ can be determined based on the target performance and the label distribution across the clients.

\subsection{FedMMB versus FedAvg}
\label{subsec:resust-vs-fed-avg}
We compare the performance of \textit{FedMMB} with \textit{FedAvg} using \textit{4CFNN} and \textit{VGG16} as models and CIFAR-10 and  HAM10000 as datasets (Figure \ref{fig:fedmmb-vs-fedavg} and Table \ref{tab:fedmmb-vs-fedavg}). We first train \textit{4CFNN} on CIFAR-10 in a federated configuration with $K=10$ clients, batch size $B=10$, and the \textit{Non-IID-2} scenario using \textit{FedMMB} ($C=20$, $\eta=0.05$) and \textit{FedAvg ($E=1$, $\eta=0.02$)}. We use a lower learning rate for \textit{FedAvg} because the model diverges for the higher learning rates.

\textit{FedMMB} and \textit{FedAvg} achieve the maximum accuracy of $0.6906$ and $0.6564$, respectively, indicating that \textit{FedMMB} outperforms \textit{FedAvg} in terms of accuracy in the \textit{Non-IID} scenario (Table \ref{tab:cifar-fedmmb-vs-fedavg}). These results are consistent with those from subsection \ref{subsec:result-fed-mmb} regarding the relationship between the number of local updates and the maximum achievable accuracy in the severely \textit{Non-IID} label distribution case assuming the same batch size. With batch size of $10$, \textit{FedMMB} and \textit{FedAvg} client $j$ performs $\mu_{j} = 20$ and $\mu_{j}=\frac{10000}{10} = 1000$ local updates per iteration, respectively ($10000$ is the sample size of each client). The approach with a lower number of local updates reaches a higher accuracy. 

We test \textit{FedAvg} with larger batch sizes of $B=100$ and $B=500$ ($E=1$, $\eta=0.05$) to perform fewer ($\mu_{j}=100$ and $\mu_{j}=20$) local updates per iteration (Figure \ref{fig:fedmmb-vs-fedavg}a and Table \ref{tab:cifar-fedmmb-vs-fedavg}). \textit{FedAvg} reaches the maximum accuracy of $0.6663$ and $0.6654$ for $B=100$ and $B=500$, respectively, which is a small improvement over \textit{FedAvg} with batch size $B=10$ ($0.6564$). Comparing the accuracy of \textit{FedMMB} ($C=20$ and $B=10$) to \textit{FedAvg} with $B=500$ ($\approx0.6906$ vs. $\approx0.6654$) highlights the importance of decoupling the batch size from the batch count (the main idea behind \textit{FedMMB}). While both approaches perform the same number of local updates on the clients ($\mu_{j}=20$),  \textit{FedMMB} achieves better accuracy because it employs a smaller batch size without affecting the batch count, which is not possible in \textit{FedAvg}.

\begin{figure}
    \centering
            \includegraphics[width=0.49\textwidth]{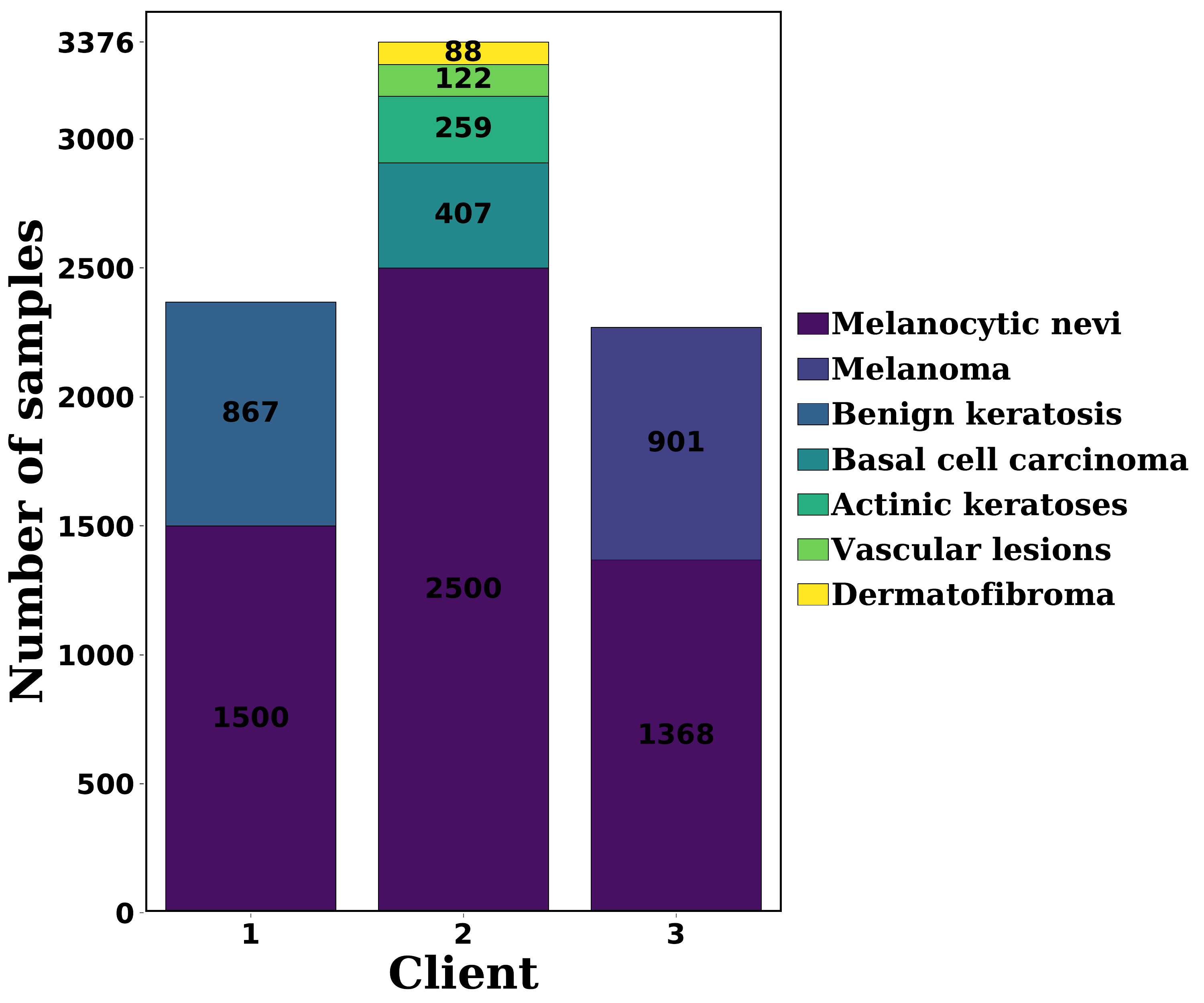}
\caption{\textit{\textit{HAM-Non-IID} scenario}}
\label{fig:ham-noniid}
\end{figure}

We also train \textit{VGG16} on the HAM10000 dataset to evaluate the performance of \textit{FedMMB} and \textit{FedAvg} on a deeper neural network and a real-world, imbalanced dataset. We use the same batch size ($B=25$) and learning rate ($\eta=0.001$) for both approaches. The batch count is $20$ for \textit{FedMMB}, while the number of local epochs is $1$ in \textit{FedAvg}. We randomly split the dataset into the train set ($8012$ images) and the test set ($2003$ images). For the \textit{Non-IID} scenario, we partition the train set among three clients ($2367$ samples of two classes, $3376$ sample from five classes, and $2269$ images from two classes) (Figure \ref{fig:ham-noniid}). Notice that class \textit{Melanocytic nevi} is still the majority class in all clients and sample distribution is imbalanced across the clients. We refer to this scenario as \textit{HAM-Non-IID}. We use \textit{AUC} (Area Under the receiver operating characteristic Curve), a common performance metric for classification tasks on imbalanced datasets, to compare the performance of the approaches in the \textit{HAM-Non-IID} scenario.

According to Figure \ref{fig:fedmmb-vs-fedavg}b and Table \ref{tab:ham-fedmmb-vs-fedavg}, \textit{FedMMB} reaches higher \textit{AUC} value than \textit{FedAvg} in the \textit{HAM-Non-IID} scenario (maximum \textit{AUC} of $0.7431$ versus $0.6931$ ). Similar to the \textit{4CFNN}-CIFAR-10 case, the large number of local updates in the \textit{FedAVG} clients adversely affects the performance in the \textit{Non-IID} setting. These results emphasize the importance of controlling the local updates on the clients, the capability that \textit{FedMMB} offers through the batch count hyperparameter. Given that, \textit{FedMMB} is a flexible approach that can provide desirable performance or communication efficiency in the \textit{Non-IID} environments with various degree of (label) heterogeneity.

%% file: conclusion.tex
In this paper, we address two main challenges of the federated learning in \textit{Non-IID} environments: performance and network communication efficiency. With respect to the performance challenge, we introduce the \textit{federated-centralized concordance} property and show that the \textit{FedSMB} approach can train federated models that are concordant with the corresponding centralized models, and therefore, it can achieve comparable performance in the \textit{Non-IID} environments and has the potential to overcome the performance challenge in the \textit{Non-IID} settings.

We also present \textit{FedMMB} as a generalization of \textit{FedSMB} to tackle the communication efficiency challenge. Unlike \textit{FedAvg}, \textit{FedMMB} decouples the batch size from the batch count and controls the number of local updates per iteration separate from the batch size. This decoupling enables \textit{FedMMB} to provide a trade-off between the performance and communication efficiency. The simulation results indicate that \textit{FedMMB} outperforms \textit{FedAvg} in terms of the accuracy and \textit{AUC} and it is a suitable training approach to federated learning in \textit{Non-IID} environments.

%% file: ms.bbl
\begin{thebibliography}{29}
\providecommand{\natexlab}[1]{#1}
\providecommand{\url}[1]{\texttt{#1}}
\expandafter\ifx\csname urlstyle\endcsname\relax
  \providecommand{\doi}[1]{doi: #1}\else
  \providecommand{\doi}{doi: \begingroup \urlstyle{rm}\Url}\fi

\bibitem[Abadi et~al.(2016)Abadi, Barham, Chen, Chen, Davis, Dean, Devin,
  Ghemawat, Irving, Isard, et~al.]{tensorflow}
Abadi, M., Barham, P., Chen, J., Chen, Z., Davis, A., Dean, J., Devin, M.,
  Ghemawat, S., Irving, G., Isard, M., et~al.
\newblock Tensorflow: A system for large-scale machine learning.
\newblock In \emph{12th $\{$USENIX$\}$ symposium on operating systems design
  and implementation ($\{$OSDI$\}$ 16)}, pp.\  265--283, 2016.

\bibitem[Bottou(2012)]{stochastic-gradient-descent}
Bottou, L.
\newblock Stochastic gradient descent tricks.
\newblock In \emph{Neural networks: Tricks of the trade}, pp.\  421--436.
  Springer, 2012.

\bibitem[Briggs et~al.(2020)Briggs, Fan, and Andras]{briggs2020federated}
Briggs, C., Fan, Z., and Andras, P.
\newblock Federated learning with hierarchical clustering of local updates to
  improve training on non-iid data.
\newblock \emph{arXiv preprint arXiv:2004.11791}, 2020.

\bibitem[Brisimi et~al.(2018)Brisimi, Chen, Mela, Olshevsky, Paschalidis, and
  Shi]{brisimi2018federated}
Brisimi, T.~S., Chen, R., Mela, T., Olshevsky, A., Paschalidis, I.~C., and Shi,
  W.
\newblock Federated learning of predictive models from federated electronic
  health records.
\newblock \emph{International journal of medical informatics}, 112:\penalty0
  59--67, 2018.

\bibitem[Chen et~al.(2020)Chen, Qin, Wang, Yu, and Gao]{chen2020fedhealth}
Chen, Y., Qin, X., Wang, J., Yu, C., and Gao, W.
\newblock Fedhealth: A federated transfer learning framework for wearable
  healthcare.
\newblock \emph{IEEE Intelligent Systems}, 2020.

\bibitem[Chollet et~al.(2021)]{keras}
Chollet, F. et~al.
\newblock Keras.
\newblock \url{https://keras.io}, 2021.

\bibitem[Hard et~al.(2018)Hard, Rao, Mathews, Ramaswamy, Beaufays, Augenstein,
  Eichner, Kiddon, and Ramage]{hard2018federated}
Hard, A., Rao, K., Mathews, R., Ramaswamy, S., Beaufays, F., Augenstein, S.,
  Eichner, H., Kiddon, C., and Ramage, D.
\newblock Federated learning for mobile keyboard prediction.
\newblock \emph{arXiv preprint arXiv:1811.03604}, 2018.

\bibitem[Hinton et~al.(2012)Hinton, Srivastava, and
  Swersky]{gradient-descent-slides}
Hinton, G., Srivastava, N., and Swersky, K.
\newblock Neural networks for machine learning lecture 6a overview of
  mini-batch gradient descent.
\newblock \emph{Cited on}, 14\penalty0 (8), 2012.

\bibitem[Hsieh et~al.(2019)Hsieh, Phanishayee, Mutlu, and
  Gibbons]{hsieh2019non}
Hsieh, K., Phanishayee, A., Mutlu, O., and Gibbons, P.~B.
\newblock The non-iid data quagmire of decentralized machine learning.
\newblock \emph{arXiv preprint arXiv:1910.00189}, 2019.

\bibitem[Jeong et~al.(2018)Jeong, Oh, Kim, Park, Bennis, and
  Kim]{jeong2018communication}
Jeong, E., Oh, S., Kim, H., Park, J., Bennis, M., and Kim, S.
\newblock Communication-efficient on-device machine learning: Federated
  distillation and augmentation under non-iid private data.
\newblock \emph{arXiv preprint arXiv:1811.11479}, 2018.

\bibitem[Kairouz et~al.(2019)Kairouz, McMahan, Avent, Bellet, Bennis, Bhagoji,
  Bonawitz, Charles, Cormode, Cummings, et~al.]{kairouz2019federated3}
Kairouz, P., McMahan, H.~B., Avent, B., Bellet, A., Bennis, M., Bhagoji, A.~N.,
  Bonawitz, K., Charles, Z., Cormode, G., Cummings, R., et~al.
\newblock Advances and open problems in federated learning.
\newblock \emph{arXiv preprint arXiv:1912.04977}, 2019.

\bibitem[Kone{\v{c}}n{\`y} et~al.(2015)Kone{\v{c}}n{\`y}, McMahan, and
  Ramage]{konevcny2015federated}
Kone{\v{c}}n{\`y}, J., McMahan, B., and Ramage, D.
\newblock Federated optimization: Distributed optimization beyond the
  datacenter.
\newblock \emph{arXiv preprint arXiv:1511.03575}, 2015.

\bibitem[Kone{\v{c}}n{\`y} et~al.(2016)Kone{\v{c}}n{\`y}, McMahan, Yu,
  Richt{\'a}rik, Suresh, and Bacon]{konevcny2016federated}
Kone{\v{c}}n{\`y}, J., McMahan, H.~B., Yu, F.~X., Richt{\'a}rik, P., Suresh,
  A.~T., and Bacon, D.
\newblock Federated learning: Strategies for improving communication
  efficiency.
\newblock \emph{arXiv preprint arXiv:1610.05492}, 2016.

\bibitem[Krizhevsky et~al.(2009)Krizhevsky, Hinton, et~al.]{cifar10}
Krizhevsky, A., Hinton, G., et~al.
\newblock Learning multiple layers of features from tiny images.
\newblock 2009.

\bibitem[Langley(2000)]{langley00}
Langley, P.
\newblock Crafting papers on machine learning.
\newblock In Langley, P. (ed.), \emph{Proceedings of the 17th International
  Conference on Machine Learning (ICML 2000)}, pp.\  1207--1216, Stanford, CA,
  2000. Morgan Kaufmann.

\bibitem[LeCun et~al.(2010)LeCun, Cortes, and Burges]{mnist}
LeCun, Y., Cortes, C., and Burges, C.
\newblock Mnist handwritten digit database.
\newblock \emph{ATT Labs [Online]. Available:
  http://yann.lecun.com/exdb/mnist}, 2, 2010.

\bibitem[Li et~al.(2020)Li, Sahu, Zaheer, Sanjabi, Talwalkar, and
  Smith]{li2020federated}
Li, T., Sahu, A.~K., Zaheer, M., Sanjabi, M., Talwalkar, A., and Smith, V.
\newblock Federated optimization in heterogeneous networks.
\newblock \emph{Proceedings of Machine Learning and Systems}, 2:\penalty0
  429--450, 2020.

\bibitem[Li et~al.(2019)Li, Huang, Yang, Wang, and Zhang]{li2019convergence}
Li, X., Huang, K., Yang, W., Wang, S., and Zhang, Z.
\newblock On the convergence of fedavg on non-iid data.
\newblock In \emph{International Conference on Learning Representations}, 2019.

\bibitem[McMahan et~al.(2017)McMahan, Moore, Ramage, Hampson, and
  Arcas]{mcmahan2017communication}
McMahan, H.~B., Moore, E., Ramage, D., Hampson, S., and Arcas, B.~A.
\newblock Communication-efficient learning of deep networks from decentralized
  data.
\newblock In \emph{Artificial Intelligence and Statistics}, pp.\  1273--1282.
  PMLR, 2017.

\bibitem[Ruder(2016)]{gradient-descent-overview}
Ruder, S.
\newblock An overview of gradient descent optimization algorithms.
\newblock \emph{arXiv preprint arXiv:1609.04747}, 2016.

\bibitem[Sattler et~al.(2019)Sattler, Wiedemann, M{\"u}ller, and
  Samek]{sattler2019robust}
Sattler, F., Wiedemann, S., M{\"u}ller, K.~R., and Samek, W.
\newblock Robust and communication-efficient federated learning from non-iid
  data.
\newblock \emph{IEEE transactions on neural networks and learning systems},
  2019.

\bibitem[Sheller et~al.(2018)Sheller, Reina, Edwards, Martin, and
  Bakas]{sheller2018multi}
Sheller, M.~J., Reina, G.~A., Edwards, B., Martin, J., and Bakas, S.
\newblock Multi-institutional deep learning modeling without sharing patient
  data: A feasibility study on brain tumor segmentation.
\newblock In \emph{International MICCAI Brainlesion Workshop}, pp.\  92--104.
  Springer, 2018.

\bibitem[Simonyan \& Zisserman(2015)Simonyan and Zisserman]{vgg16}
Simonyan, K. and Zisserman, A.
\newblock Very deep convolutional networks for large-scale image recognition.
\newblock In \emph{International Conference on Learning Representations}, 2015.

\bibitem[Tschandl et~al.(2018)Tschandl, Rosendahl, and Kittler]{ham10000}
Tschandl, P., Rosendahl, C., and Kittler, H.
\newblock The ham10000 dataset, a large collection of multi-source
  dermatoscopic images of common pigmented skin lesions.
\newblock \emph{Scientific data}, 5\penalty0 (1):\penalty0 1--9, 2018.

\bibitem[Wang et~al.(2020{\natexlab{a}})Wang, Kaplan, Niu, and
  Li]{wang2020optimizing}
Wang, H., Kaplan, Z., Niu, D., and Li, B.
\newblock Optimizing federated learning on non-iid data with reinforcement
  learning.
\newblock In \emph{IEEE INFOCOM 2020-IEEE Conference on Computer
  Communications}, pp.\  1698--1707. IEEE, 2020{\natexlab{a}}.

\bibitem[Wang et~al.(2020{\natexlab{b}})Wang, Yurochkin, Sun, Papailiopoulos,
  and Khazaeni]{wang2020federated}
Wang, H., Yurochkin, M., Sun, Y., Papailiopoulos, D., and Khazaeni, Y.
\newblock Federated learning with matched averaging.
\newblock \emph{arXiv preprint arXiv:2002.06440}, 2020{\natexlab{b}}.

\bibitem[Xiao et~al.(2017)Xiao, Rasul, and Vollgraf]{fashion-mnist}
Xiao, H., Rasul, K., and Vollgraf, R.
\newblock Fashion-mnist: a novel image dataset for benchmarking machine
  learning algorithms.
\newblock \emph{arXiv preprint arXiv:1708.07747}, 2017.

\bibitem[Yang et~al.(2018)Yang, Andrew, Eichner, Sun, Li, Kong, Ramage, and
  Beaufays]{yang2018applied}
Yang, T., Andrew, G., Eichner, H., Sun, H., Li, W., Kong, N., Ramage, D., and
  Beaufays, F.
\newblock Applied federated learning: Improving google keyboard query
  suggestions.
\newblock \emph{arXiv preprint arXiv:1812.02903}, 2018.

\bibitem[Zhao et~al.(2018)Zhao, Li, Lai, Suda, Civin, and
  Chandra]{zhao2018federated}
Zhao, Y., Li, M., Lai, L., Suda, N., Civin, D., and Chandra, V.
\newblock Federated learning with non-iid data.
\newblock \emph{arXiv preprint arXiv:1806.00582}, pp.\  1--3, 2018.

\end{thebibliography}
